# Adversarial Scene Reconstruction and Object Detection System for Assisting Autonomous Vehicle

Md Foysal Haque, Hay-Youn Lim, and Dae-Seong Kang
*Department of Electronic Engineering*
*Dong-A University, Republic of Korea*
tusnow1990@gmail.com, hylim@dau.ac.kr, dskang@dau.ac.kr

**Abstract**

*In the current computer vision era classifying scenes through video surveillance systems is a crucial task. Artificial Intelligence (AI) Video Surveillance technologies have been advanced remarkably while artificial intelligence and deep learning ascended into the system. Adopting the superior compounds of deep learning visual classification methods achieved enormous accuracy in classifying visual scenes. However, the visual classifiers face difficulties examining the scenes in dark visible areas, especially during the nighttime. Also, the classifiers face difficulties in identifying the contexts of the scenes. This paper proposed a deep learning model that reconstructs dark visual scenes to clear scenes like daylight, and the method recognizes visual actions for the autonomous vehicle. The proposed model achieved 87.3 percent accuracy for scene reconstruction and 89.2 percent in scene understanding and detection tasks.*

## 1. Introduction

Recent research interest in autonomous vehicles accelerated new technologies' improvement, forming a very active research field. Problem statements in this area are often correlated to computer vision, as assuming what is progressing on outside and handling uncertain situations even under difficult weather conditions is crucial for traffic safety. This is pretty similar to the challenges visually impaired people suffer in their daily life. Therefore, the booming field of autonomous driving can also accelerate progress in the very smaller region of assistive technologies for the blind. Bringing those applications together and discussing how modern aids for visually impaired people can benefit from the enormously larger automotive industry.

Scene understanding is complex and multi-dimensional because traffic scene combines continuously with the surrounding environment and people within a dynamic process. In Figure 1 illustrates the scene classifying and understanding process of google self-driving car. Robust scene understanding deep learning frameworks [1-2] adopted semantic segmentation [3], feature extraction [4] and achieved enormous performance to solve the visual classification tasks. However, due to the complex structure and expensive computation, it is not easy to achieve higher visual task precision. The proposed model proposed to classify the visual scenes adopting improved deep learning architecture concerning the above issue.

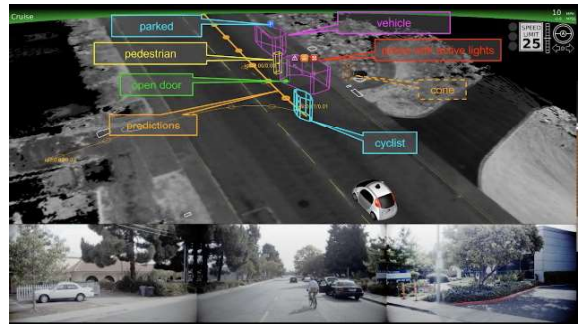

**Figure 1. Vision guiding system for google autonomous car [5].**

The proposed approaches consist of two parts, essentially generative adversarial network (GAN) [6] and object detection module [7]. Mainly two different approaches are applied for scene reconstruction through GAN and feature extraction: computing optical flow information of the output images for object detection. Furthermore, computing deep convolutional neural networks-based representations.

## 2. Related Theories

Scene understanding and action recognition is the core problem of computer vision, which involves several applications, including robotics, surveillance, and the automotive industry.

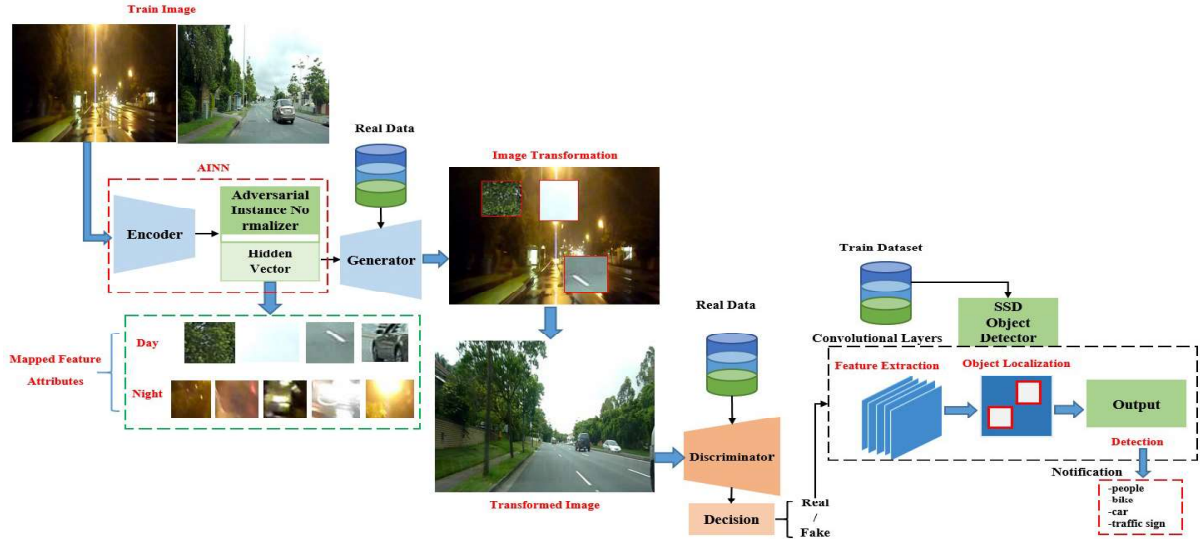

**Figure 2.** The proposed architecture of adversarial scene reconstruction and object detection system for autonomous vehicle.

The recent developments of high-performance computers became the key to process a large amount of data at high speed. The deep learning mainstream method extracts a feature vector (called the image local features) from the image and applies a learning method to perform scene understanding and action recognition.

One of the most typical computer vision solutions for scene classification and recognition is two-stream convolutional networks. In this paper, two CNNs [8] are used, one for visual reconstruction and another one adopt spatial feature extraction, which examines the actions from single images for the temporal feature extraction, which determines from the optical flow vectors of multiple frames. Then, outputs of the two networks are combined at the end with detection.

## 3. Proposed Network

Multiple algorithms formed for self-driving cars can also be used as parts of a navigation assistant for the visually impaired. Both tasks perform similar difficulties for image processing, scene understanding, obstacle recognition, outlining, localization, and object detection, among others. Considering these assisting elements of self-driving cars, a visual guiding system was proposed to assist the autonomous car vision system. The overall network architecture of the proposed approach is shown in Figure 2. The network apportions into two-part first one constructed with improved Adversarial Instance Normalization Network (AINN v2) for image scene reconstruction. It carries image-to-image scene reconstruction applying the fundamentals of domain shifting. Moreover, the network examines the source data to characterize the feature differences from the two domains' transformation tasks. The scene reconstruction task follows equation 1.

$$Domain_{(a\ to\ b)} = I_a[(D(b)-1)^2 + I_b[D(G(a))^2] \quad (1)$$

where, $I_a$ and $I_b$ are the image domains, $G$ is generator, and the discriminator is $D$.

After successfully reconstructing scenes from night to day scenes, the transformed image passed to the improved Single Shot MultiBox Detector (SSD) [7] object detection network. The module follows the localization and detection process regarding equation 2.

$$B_{box} = \sum_{pos}^{N} C\left(l(x,c) + \varphi Ground_L(x,y)\right) \quad (2)$$

where, $C$ represents the confidence map, predicted box is $l$, $Ground_L$ represents the ground truth localization, $\varphi$ is feature coefficient, and (x, y) is object area.

The network is deployed to understand the scenes from the image, examine the objects using spatial feature extraction and localization, and resolve the detection task. The detected object's knowledge is utilized to guide the mainstream of autonomous cars.

## 4. Experiments and Results

The image-to-image reconstruction model required unpaired images. The goal of the study is to transform the night images into day images and examine the image and understand the scenes to assist the self-driving car.

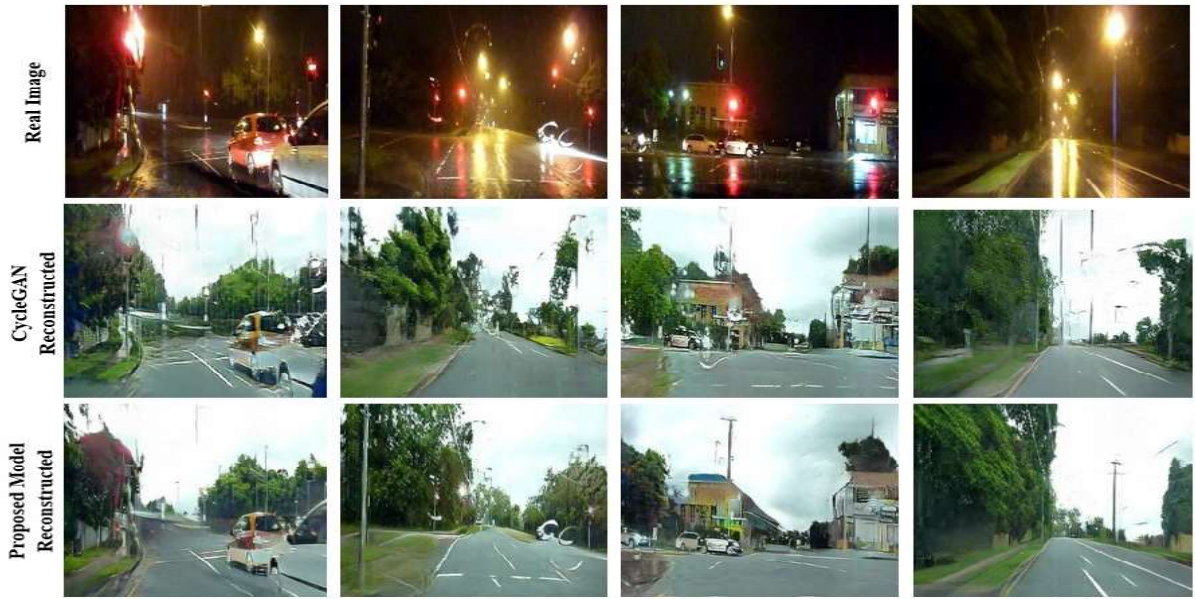

**Figure 3. Comparison of reconstructed results between proposed model and CycleGAN. Top to bottom row: real rainy night images before scene reconstruction, synthetic day images using the proposed adversarial model and CycleGAN.**

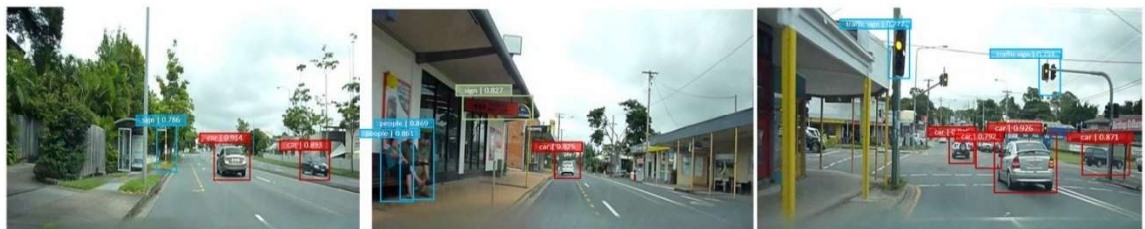

**Figure 4. Detection results of proposed model.**

**Table 1. Summary of mAP (mean-average precision) for each category, inference speed, and average detection accuracy.**

| Method | bike | bus | car | people | sign | traffic sign | FPS | Average mAP (%) |
|---|---|---|---|---|---|---|---|---|
| **Proposed model +SSD** | **89.6** | **91.4** | **89.8** | **91.7** | **84.1** | **85.7** | **33.8** | **89.2** |
| FRCN+VGG [11] | 81.3 | 83.5 | 86.0 | 87.9 | - | - | 1.7 | 74.8 |
| ResNet50-Det [2] | 80.0 | 81.3 | 86.7 | 89.3 | - | - | 22.5 | 76.1 |

**Table 2. Comparison of the results of scene reconstruction with robust adversarial models.**

| Method | Dataset | Input Size | Reconstruction Accuracy |
|---|---|---|---|
| **Proposed Model (AINN v2)** | Alderley Day/Night | 286×286 | **87.3%** |
| CycleGAN [11] | | | 79.7% |
| AINN [6] | | | 86.2% |

We used two different datasets to train those networks, the AINN v2 network trained with Alderley Day/Night dataset [9] and the SSD object detection network trained with a custom dataset that included Alderly dataset images along with the Cityscapes dataset images [10]. The evaluated results of the proposed model shown separately in two figures.

The scene reconstruction results are shown in Figure 3, and detection results are shown in Figure 4. Moreover, the evaluation matrix of adversarial scene reconstruction and overall detection ablation are shown in Table 1 and Table 2. To evaluate the proposed network's performance, the scene reconstructed results are compared with conventional CycleGAN. Considering all the reconstruction results, the proposed model showed the enhanced performance to translate rainy night scenes into clear daylight scenes. Moreover, using the reconstructed image detection task conducted and the object detection module trained with 6 different classes of objects. As same as scene reconstruction task detection task also achieved higher accuracy to detect and understand the scenes.

## 5. Conclusions

This paper proposed an image-to-image reconstruction and object detection and scene understanding framework applied to reconstruct the rainy-night image scene to add daylight scenes. The network adopts an adversarial network for visual scene reconstruction and detecting object a spatial feature extraction, and localization-based SSD model used for the system. Moreover, the framework reconstructs the images of rainy-night scenarios into daylight scenarios. The network generates the daylight image to improve visibility without harming any objects from the source image to assist the mainframe of the module in notifying the objects and scenes. The proposed network achieved 87.3 percent accuracy in the reconstruction task, and the network achieved 89.2 percent accuracy in the object detection task. In the future, we will train our proposed module to understand pedestrians' actions and traffic scenes for assisting and solving autonomous cars' vision tasks.

## Acknowledgment

This work was supported by the National Research Foundation of Korea (NRF) grant funded by the Korea government (NO.2017R1D1A1B04030870).

## References

[1] J. Choi, D. Chun, H. Kim, and H.J, Lee, "Gaussian YOLOv3: An Accurate and Fast Object Detector Using Localization Uncertainty for Autonomous Driving", In Proceedings of the IEEE/CVF International Conference on Computer Vision, pp. 502-511, 2019.

[2] B. Wu, A. Wan, F. Iandola, P. H. Jin, and K. Keutzer, "SqueezeDet: Unified, Small, Low Power Fully Convolutional Neural Networks for Real-Time Object Detection for Autonomous Driving", In Proceedings of the IEEE Conference on Computer Vision and Pattern Recognition Workshops, pp. 129-137, 2017.

[3] L.C. Chen, G. Papandreou, F. Schroff, and H. Adam, "Rethinking Atrous Convolution for Semantic Image Segmentation", arXiv preprint arXiv:1706.05587, June 2017.

[4] S. Ren, K. He, R. Girshick, and J. Sun, "Faster R-CNN: Towards Real-Time Object Detection with Region Proposal Networks", arXiv preprint arXiv:1506.01497, June 2015.

[5] J. Cohen, "How Google's Self-Driving Cars Work", August, 2020. [Online]. Available: https://heartbeat.fritz.ai/how-googles-self-driving-cars-work-c77e4126f6e7

[6] M. F. Haque, and D.S. Kang, "AINN: Adversarial Instance Normalization Network for Image-to-Image Translation", The Proceedings of KIIT Conference, vol. 15, no. 1, pp. 117-120, 2020.

[7] M. F. Haque, and D.S. Kang, "Multi Scale Object Detection Based on Single Shot Multibox Detector with Feature Fusion and Inception Network", The Journal of Korean Institute of Information Technology, vol. 16, no. 10, pp. 93-100, 2018.

[8] A. Krizhevsky, I. Sutskever, and G. E. Hinton. "ImageNet classification with deep convolutional neural networks", NIPS, pp. 1097-1105, 2012.

[9] M. J. Milford, and G. F. Wyeth, "SeqSLAM: Visual Route-Based Navigation for Sunny Summer Days and Stormy Winter Nights", In IEEE international conference on robotics and automation, pp. 1643-1649, 2012.

[10] M. Cordts, M. Omran, S. Ramos, T. Rehfeld, M. Enzweiler, R. Benenson, U. Franke, S. Roth, and B. Schiele, "The Cityscapes Dataset for Semantic Urban Scene Understanding", In Proceedings of the IEEE conference on computer vision and pattern recognition, pp. 3213-3223, 2016.

[11] K. Ashraf, B. Wu, F. N. Iandola, M. W. Moskewicz, and K. Keutzer, "Shallow networks for high-accuracy road object detection", arXiv:1606.01561, 2016.

[12] M.Y. Liu, T. Breuel, and J. Kautz, "Unsupervised Image-to-Image Translation Networks", arXiv preprint arXiv:1703.00848, 2017.